\documentclass[11pt,a4paper]{article}
\usepackage[hyperref]{acl2019}

\usepackage{times}
\usepackage{latexsym}
\usepackage{bm}
\usepackage{url}
\usepackage{graphicx}
\usepackage{amsmath}
\usepackage{multirow}

\aclfinalcopy 

\DeclareMathOperator*{\argminA}{arg\,min} 

\title{ANA at SemEval-2019 Task 3:  Contextual Emotion detection in Conversations through hierarchical LSTMs and BERT }

\author{Chenyang Huang, Amine Trabelsi, Osmar R. Za\"{\i}ane\\
  Department of Computing Science, University of Alberta \\
  {\tt \{chuang8,atrabels,zaiane\}@ualberta.ca} \\}

\date{}

\begin{document}
\maketitle

\begin{abstract}
    This paper describes the system submitted by ANA Team for the SemEval-2019 Task 3: EmoContext.
    We propose a novel Hierarchical LSTMs for Contextual Emotion Detection (HRLCE) model. 
    It classifies the emotion of an utterance given its conversational context.
    The results show that, in this task, our HRCLE outperforms the most recent state-of-the-art text classification framework: BERT. We combine the results generated by BERT and HRCLE to achieve an overall score of 0.7709 which ranked 5th on the final leader board of the competition among 165 Teams.
    
\end{abstract}

\section{Introduction}

Social media has been a fertile environment 
for the expression of opinion and emotions via text.
The manifestation of this expression differs from traditional or conventional opinion communication in text (e.g., essays). It is usually short (e.g. Twitter), containing new forms of constructs, including emojis, hashtags or slang words, etc. This constitutes a new challenge for the NLP community. 
Most of the studies in the literature focused on the detection  
of sentiments (i.e. positive, negative or neutral)  \cite{mohammad2013crowdsourcing,kiritchenko2014sentiment}.

Recently,  emotion classification from social media text started receiving more attention \cite{Mohammad-semeval18,SentimentSurvey17}.
Emotions have been extensively studied in psychology \cite{ekman1992argument,plutchik2001}. 
Their automatic detection
may reveal important information in social online environments, like
online customer service. 
In such cases, a 
user is conversing with an automatic chatbot. 
Empowering the chatbot with the ability to detect the user's emotion is a step forward towards the construction of an emotionally intelligence agent. Giving the detected emotion, an emotionally intelligent agent would generate an empathetic response. 
Although its potential convenience, detecting emotion in textual conversation has seen limited attention so far. One of the main challenges is that one user’s utterance may be insufficient to recognize the emotion~\cite{NAACL18}.
The need to consider the context of the conversion is essential in this case, even for human, specifically given the lack of voice modulation and facial expressions. The usage of figurative language, like sarcasm, and the class size's imbalance adds up to this problematic \cite{chatterjee2019understanding}. 



\begin{figure}[h!]
    \centering
    \includegraphics[keepaspectratio=true, width=0.48\textwidth]{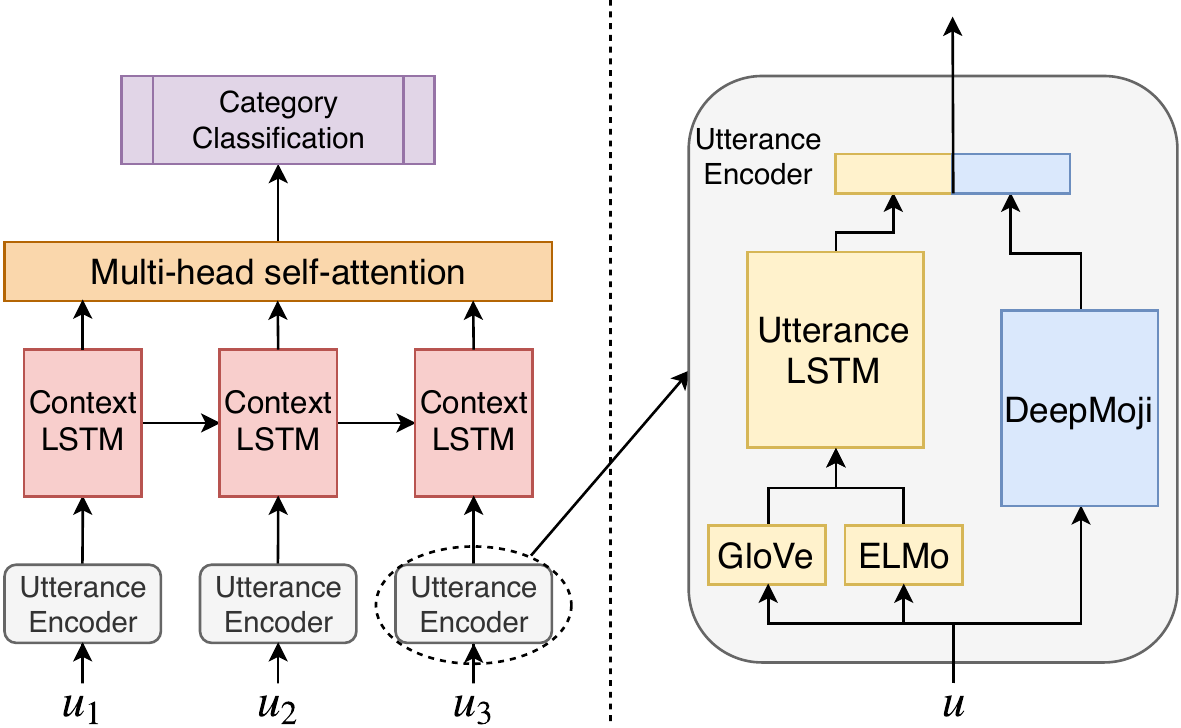}
    \caption[An illustration of the HRLCE model] {An illustration of the HRLCE model}
    \label{fig:high_level_model}
\end{figure}

In this paper, we describe our model, which was proposed for the SemEval 2019-Task 3 competition: Contextual Emotion Detection in Text (EmoContext). The competition consists in classifying the emotion of an utterance given its conversational context. More formally, given a textual user utterance along with 2 turns of context in a conversation, the task is to classify the emotion of user utterance as Happy, Sad, Angry or Others \cite{SemEval2019Task3}. The conversations are extracted from Twitter.

We propose an ensemble approach composed of two 
deep learning models, 
the \emph{Hierarchical LSTMs for Contextual Emotion Detection} (HRLCE) model and the BERT model \cite{devlin2018bert}. The BERT is a pre-trained language model that has shown great success in many NLP classification tasks. Our main contribution consists in devising the HRLCE model.


Figure \ref{fig:high_level_model} illustrates the main components of 
the HRLCE model.
We examine a transfer learning approach with several pre-trained models in order to encode each user utterance 
semantically and emotionally at the word-level. 
The proposed model uses Hierarchical LSTMs \cite{sordoni2015hierarchical} followed by a multi-head self attention mechanism \cite{vaswani2017attention} 
for a contextual encoding at the utterances level.

 
The model evaluation on the competition's test set resulted in a 0.7709 harmonic mean of the macro-F1 scores across the categories \emph{Happy}, \emph{Angry}, and \emph{Sad}. This result ranked 5th in the final leader board of the competition among 142 teams 
with a score above 
the organizers' baseline.





\section{Overview}
\subsection{Embeddings for semantics and emotion}
We use different kinds of embeddings that have been deemed effective in the literature in capturing not only the syntactic or semantic information of the words, but also their emotional content. We breifly describe them in this section.

GloVe, \cite{pennington2014glove} is a widely used pre-trained vector representation that captures fine-grained syntactic and semantic regularities. It has shown great success in 
word similarity tasks and Named Entity Recognition benchmarks. 

ELMo, or Embeddings from Language Models, \cite{peters2018deep} are  deep contextualized word representations. These representations enclose a polysemy encoding, i.e., they capture the variation
in the meaning of a word 
depending on
its context.
The representations are learned functions of the input, pre-trained with deep bi-directional LSTM model.
It has been shown to work well in practice on multiple language understanding tasks like question answering, entailment and sentiment analysis.
In this work, our objective is to detect emotion accurately giving the context. Hence, employing such contextual embedding can be crucial.




DeepMoji \cite{felbo2017} is a pre-trained model containing rich representations of emotional content. It has been pre-trained on the task of predicting the emoji contained in the text using Bi-directional LSTM layers combined with an attention layer. A distant supervision approach was deployed to collect a massive (1.2 billion Tweets) dataset with diverse set of noisy emoji labels on which DeepMoji is pre-trained. 
This led to state-of-the art performance when fine-tuning DeepMoji on a range of target tasks related to sentiment, emotion and sarcasm.

\subsection{Hierarchical RNN for context}
One of the building component of our proposed model (see Figure \ref{fig:high_level_model}) is the Hierarchical or Context
recurrent encoder-decoder (HRED) \cite{sordoni2015hierarchical}. 
HRED architecture  is used for encoding dialogue context in the task of multi-turn dialogue generation task \cite{serban2016building}. It has been proven to be effective in capturing the context information of dialogue exchanges. It contains two types of recurrent neural net (RNN) units: \emph{encoder} RNN which maps each utterance to an utterance vector; \emph{context} RNN which further processes the utterance vectors. HRED is expected to 
produce a better representation of the context in  
dialogues because the \emph{context} RNN allows the model to represent the information exchanges between the two speakers.

\subsection{BERT}
\label{sec:BERT}
BERT, the Bidirectional Encoder Representations for Transformers, \cite{devlin2018bert} is a pre-trained model producing context representations that can be very convenient and effective. BERT representations can be fine-tuned to many downstream NLP tasks by adding just one additional output layer for the target task, eliminating the need for engineering a specific architecture for a task. Using this setting, it has advanced the state-of-the-art performances in 11 NLP tasks. 
Using BERT in this work has slightly improved the final result, when we combine it with our HRLCE in an ensemble setting.


\subsection{Importance Weighting}
Importance Weighting \cite{sugiyama2012machine} is used when label distributions between the training and test sets are generally different, which is the case of the competition datasets (Table \ref{tab:data_dist}).
It corresponds to weighting the samples according to their importance when calculating the loss.  

A supervised deep learning model can be regarded as a parameterized function $f(\bm{x};\bm{\theta})$. The backpropagation learning algorithm through a differentiable loss is a method of \emph{empirical risk minimization} (ERM). Denote $(\bm{x}_i^{tr}, y_i^{tr})$, $i \in [1 \dots n_{tr}]$ are pairs of training samples, testing samples are $(\bm{x}_i^{te}, y_i^{te})$, $i \in [1 \dots n_{te}]$.

The ratio of $P_{te}(\bm{x}_i^{tr}) / P_{tr}(\bm{x}_i^{tr})$ is referred as the \emph{importance} of a traning sample $\bm{x}_i^{tr}$.
When the probability of an input $\bm{x}_i^{tr}$ in training and testing sets are generally different: $P_{tr}(\bm{x}_i^{tr}) \ne P_{te}(\bm{x}_i^{tr})$, the training of the model $f_{\bm{\theta}}$ is then called under \emph{covariate shift}. In such situation, the parameter $\hat{\bm{\theta}}$ should be estimated through \emph{importance-weighted ERM}:
\begin{align}
\label{eq:erm}
    \argminA_{\bm{\theta}} \Big[ \frac{1}{n_{tr}} \sum_{i=1}^{n_{tr}} \frac{P_{te}(\bm{x}_i^{tr})}{P_{tr}(\bm{x}_i^{tr})}\text{loss} (y_i^{tr}, f(\bm{x}_i^{tr}; \bm{\theta} ) \Big].
\end{align}

\section{Models}
\label{sec:models}
Denote the input $\bm{x} = [u_1, u_2, u_3]$, where $u_i$ is the $i$th penultimate utterance in the dialogue. $y$ is the emotion expressed in $u_3$ while giving $u_1$ and $u_2$ as context.

To justify the effectiveness of the modules in HRLCE, we propose two baseline models: SA-LSTM (SL) and SA-LSTM-DeepMoji (SLD). The SL model is part of the SLD model. The latter one corresponds to the utterance encoder of our HRLCE. Therefore, we illustrate the models consecutively in Sections \ref{sec:sl}, \ref{sec:sld}, and \ref{sec:hrlce}. 

\subsection{SA-LSTM (SL)}
\label{sec:sl}
Let $\bm{x}$ be the concatenation of $u_1$ ,$u_2$, and $u_3$. Hereby, $\bm{x} = [x_1, x_2, \cdots, x_n]$, where $x_i$ is the $i$th word in the combined sequence.  Denote the pre-trained GloVe model as $G$. As GloVe  model can be directly used by looking up the word $x_i$, we can use $G(x_i)$ to represent its output. 
On the contrary,
ELMo embedding 
is not just dependent on the word $x_i$, but on all the words of the input sequence.
When taking as input the entire sequence $\bm{x}$, $n$ vectors can be extracted from the pre-trained ElMo model. Denote the vectors as $\bm{E} = [E_1, E_2, \cdots, E_n]$. $E_i$ contains both contextual and semantic information of word $x_i$. We use a two-layer bidirectional LSTM as the encoder of the sequence $\bm{x}$. For simplicity, we denote it as $LSTM^e$. 
In order to better represent the information of $x_i$, we use the concatenation of $G(x_i)$ and $E_i$ as the feature embedding of $x_i$. Therefore, we have the following recurrent progress:
\begin{equation}
    h^{e}_{t}  = LSTM^e([G(x_{t}); E_{t}], h^{e}_{t-1}).
\end{equation}
$h_{t}^{e}$ is the hidden state of encoder LSTM at time step $t$, and $h_0^{e} = \bm{0}$. Let $\bm{h}^{e}_{\bm{x}} = [h^{e}_{1}, h^{e}_{2}, \cdots, h^{e}_{n}]$ be the $n$ hidden states of encoder given the input $\bm{x}$. Self-attention mechanism has been proven to be effective in helping RNN dealing with dependency problems \cite{lin2017structured}. We use the multi-head version of the self-attention~\cite{vaswani2017attention} and set the number of channels for each head as 1. Denote the self-attention module as $SA$, it takes as input all the hidden states of the LSTM and summarizes them into a single vector. This process is represented as $h^{sa}_{\bm{x}} = SA(\bm{h}^{e}_{\bm{x}})$. To predict the model, we append a fully connected (FC) layer to project $h^{sa}_{\bm{x}}$ on to the space of emotions. Denote the FC layer as $output$. Let $o^{SL}_{\bm{x}} = output(h^{sa}_{\bm{x}})$, then the estimated label of $\bm{x}$ is 
the
$\operatorname*{arg\,max}_i (o^{SL}_{\bm{x}})$, where $i$ is $i$th value in the vector $o^{SL}_{\bm{x}}$.

\subsection{SA-LSTM-DeepMoji (SLD)} 
\label{sec:sld}
SLD is the combination of SL and DeepMoji. An SLD model without the output layer and the self-attention layer is in fact the utterance encoder of the proposed HRLCE, which is illustrated in the right side of Figure \ref{fig:high_level_model}. Denote the DeepMoji model as $D$, when taking as input $\bm{x}$, the output is represented as $h^{d}_{\bm{x}} = D(\bm{x})$. We concatenate $h^{d}_{\bm{x}}$ and $h^{sa}_{\bm{x}}$ as the feature representation of sequence of ${\bm{x}}$. Same as SL, an FC layer is added in order to predict the label: $o^{SLD}_{\bm{x}} = output([h^{sa}_{\bm{x}}; h^{d}_{\bm{x}}])$.

\begin{table}[]
\centering
	\resizebox{0.48\textwidth}{!}{
        \begin{tabular}{l|l|lll|c} \hline\hline
                               & F1   & Happy  & Angry  & Sad    & Harm. Mean \\ \hline
        \multirow{2}{*}{SL}    & Dev  & 0.6430 & 0.7530 & 0.7180 & 0.7016   \\
                               & Test & 0.6400 & 0.7190 & 0.7300 & 0.6939    \\ \hline
        \multirow{2}{*}{SLD}   & Dev  & 0.6470 & 0.7610 & 0.7360 & 0.7112    \\
                               & Test & 0.6350 & 0.7180 & 0.7360 & 0.6934    \\ \hline
        \multirow{2}{*}{HRLCE} & Dev  & 0.7460 & 0.7590 & 0.8100 & \textbf{0.7706}    \\
                               & Test & 0.7220 & 0.766  & 0.8180 & \textbf{0.7666}    \\ \hline
        \multirow{2}{*}{BERT}  & Dev  & 0.7138 & 0.7736 & 0.8106 & 0.7638    \\
                               & Test & 0.7151 & 0.7654 & 0.8157 & 0.7631    \\ \hline \hline   
        \end{tabular}
}
\caption{Macro-F1 scores and its harmonic means of the four models}
\label{tab:results}
\end{table}

\subsection{HRLCE}
\label{sec:hrlce}
Unlike SL and SLD, the input of HRLCE is not the concatenation of $u_1$, $u_2$, and $u_3$. 

Following the annotation in Section \ref{sec:sl} and \ref{sec:sld}, an utterance $u_i$ is firstly encoded as $h^{e}_{u_i}$ and $h^{d}_{u_i}$. We use another two layer bidirectional LSTM as the context RNN, denoted as $LSTM^c$. Its hidden states are iterated through:

\begin{equation}
    h^{c}_{t}  = LSTM^c([h^{e}_{u_t}; h^{d}_{u_t}], h^{c}_{t-1}),
\end{equation}

where $h^{c}_{0} = \bm{0}$. The three hidden states $\bm{h^{c}} = [h^{c}_1, h^{c}_2, h^{c}_3]$, are fed as the input to a self-attention layer. The resulting vector $SA(\bm{h^{c}})$ is also projected to the label space by an FC layer.

\subsection{BERT} 
BERT (Section \ref{sec:BERT}) can take as input either a single sentence or a pair of sentences. A ``sentence'' here corresponds to any arbitrary span of contiguous words.
In this work, in order to fine-tune 
BERT, 
we concatenate utterances $u_1$ and $u_2$ to constitute 
the first sentence of the pair.
$u_3$ 
is 
the second sentence of the pair.
The reason behind such setting is that we assume that the target emotion 
$y$ is directly related to $u_3$, while $u_1$ and $u_2$ are providing additional context information. 
This forces the model to consider $u_3$ differently. 

\section{Experiment}

\subsection{Data preprocessing}
From the training data we notice that emojis are playing an important role in expressing emotions. We first use \emph{ekphrasis} package  \cite{baziotis-pelekis-doulkeridis:2017:SemEval2} to clean up the utterances. \emph{ekphrasis} 
corrects misspellings, handles textual emotions (e.g. `:)))'), and normalizes tokens (hashtags, numbers, user mentions etc.). In order to keep the semantic meanings of the emojis, we use the \emph{emojis} package\footnote{https://pypi.org/project/emoji/} to first convert them into their textual aliases and then replace the ``:'' and ``\_'' with spaces. 

\subsection{Environment and hyper-parameters}
We use PyTorch 1.0 for the deep learning framework, and our code in Python 3.6 can be accessed in GitHub\footnote{https://github.com/chenyangh/SemEval2019Task3}.
For fair comparisons, we use the same parameter settings for the common modules that are shared by the SL, SLD, and HRLCE. The dimension of \emph{encoder} LSTM is set to 1500 per direction; the dimension of \emph{context} LSTM is set to 800 per direction. We use Adam optimizer with initial learning rate as 5e-4 and a decay ratio of 0.2 after each epoch. 
The parameters of DeepMoji are set to trainable. We use \textit{BERT-Large} pre-trained model which contains 24 layers.

\begin{table}[h]
    \centering
    \setlength\tabcolsep{2.7pt}
    \begin{tabular}{l|llll|l} \hline  \hline
              & happy   & angry   & sad     & others   & size   \\  \hline
        Train & 14.07\% & 18.26\% & 18.11\% & 49.56\%  & 30160  \\
        Dev   & 5.15\%  & 5.44\%  & 4.54\%  & 84.86\%  & 2755 \\
        Test  & 4.28\%  & 5.57\%  & 4.45\%  & 85.70\%  & 5509 \\  \hline \hline
    \end{tabular}
    \caption{Label distribution of train, dev, and test set}
    \label{tab:data_dist}
\end{table}

According to the description in \cite{competition}, the label distribution for \textit{dev} and \textit{test} sets are roughly 4\% for each of the emotions. 
However, from the \textit{dev} set (Table~\ref{tab:data_dist}) we know that the proportions of each of the emotion categories are better described as \%5 each, thereby we use \%5 as the empirical estimation of distribution $P_{te}(\bm{x}_{i}^{tr})$. We did not use the exact proportion of \textit{dev} set as the estimation to prevent the overfitting towards \textit{dev} set. The sample distribution of the \textit{train} set is used as $P_{tr}(\bm{x}_i^{tr})$. We use \emph{Cross Entropy} loss for all the aforementioned models, and the loss of the training samples are weighted according to Eq.~\ref{eq:erm}.


\subsection{Results and analysis}
We run 9-fold cross validation on the training set. Each iteration, 1 fold is used to prevent the models from overfitting while the remaining folds are used for training. Therefore, every model is trained 9 times to ensure  stability. The inferences over \emph{dev} and \emph{test} sets are performed on each iteration. We use the majority voting strategy to merge the results from the 9 iterations. The results are shown in Table~\ref{tab:results}. It shows that the proposed HRLCE model performs the best.
The performance of SLD and SL are very close to each other, on the \emph{dev} set, SLD performs better than SL but they have almost the same overall scores on the \emph{test} set. The Macro-F1 scores of each emotion category are very different from each other: the classification accuracy for emotion \emph{Sad} is the highest in most of the cases, while the emotion \emph{Happy} is the least accurately classified by all the models. We also noticed that the performance on the \emph{dev} set is generally slightly better than that on the \emph{test} set. 

\section{Conclusions}
Considering the competitive results generated by BERT, we combined BERT and our proposed model in an ensemble and obtained 0.7709 on the final test leaderboard.
From a confusion matrix of our final submission, we notice that there are barely miss-classifications among the three categories  (\emph{Angry, Sad}, and \emph{Happy}). For example, the emotion Sad is rarely miss-classified as ``Happy'' or ``Angry".  
Most of the errors correspond to classifying the emotional utterances in the \emph{Others} category.
We think, as future improvement, the models need to first focus on the binary classification ``Others'' versus ``Not-Others'', then the ``Not-Others'' are classified in their respective emotion. 

\newpage

\bibliography{acl2019}

\begin{thebibliography}{20}
\expandafter\ifx\csname natexlab\endcsname\relax\def\natexlab#1{#1}\fi

\bibitem[{Baziotis et~al.(2017)Baziotis, Pelekis, and
  Doulkeridis}]{baziotis-pelekis-doulkeridis:2017:SemEval2}
Christos Baziotis, Nikos Pelekis, and Christos Doulkeridis. 2017.
\newblock Datastories at semeval-2017 task 4: Deep lstm with attention for
  message-level and topic-based sentiment analysis.
\newblock In \emph{Proceedings of the 11th International Workshop on Semantic
  Evaluation (SemEval-2017)}, pages 747--754, Vancouver, Canada. Association
  for Computational Linguistics.

\bibitem[{Chatterjee et~al.(2019{\natexlab{a}})Chatterjee, Gupta, Chinnakotla,
  Srikanth, Galley, and Agrawal}]{chatterjee2019understanding}
Ankush Chatterjee, Umang Gupta, Manoj~Kumar Chinnakotla, Radhakrishnan
  Srikanth, Michel Galley, and Puneet Agrawal. 2019{\natexlab{a}}.
\newblock Understanding emotions in text using deep learning and big data.
\newblock \emph{Computers in Human Behavior}, 93:309--317.

\bibitem[{Chatterjee et~al.(2019{\natexlab{b}})Chatterjee, Narahari, Joshi, and
  Agrawal}]{SemEval2019Task3}
Ankush Chatterjee, Kedhar~Nath Narahari, Meghana Joshi, and Puneet Agrawal.
  2019{\natexlab{b}}.
\newblock Semeval-2019 task 3: Emocontext: Contextual emotion detection in
  text.
\newblock In \emph{Proceedings of The 13th International Workshop on Semantic
  Evaluation (SemEval-2019)}, Minneapolis, Minnesota.

\bibitem[{CodaLab(2019)}]{competition}
CodaLab. 2019.
\newblock Semeval19 task 3: Emocontext.
\newblock
  \url{https://competitions.codalab.org/competitions/19790\#learn\_the_details-data-set-format}.

\bibitem[{Devlin et~al.(2018)Devlin, Chang, Lee, and
  Toutanova}]{devlin2018bert}
Jacob Devlin, Ming-Wei Chang, Kenton Lee, and Kristina Toutanova. 2018.
\newblock Bert: Pre-training of deep bidirectional transformers for language
  understanding.
\newblock \emph{arXiv preprint arXiv:1810.04805}.

\bibitem[{Ekman(1992)}]{ekman1992argument}
Paul Ekman. 1992.
\newblock An argument for basic emotions.
\newblock \emph{Cognition \& emotion}, 6(3-4):169--200.

\bibitem[{Felbo et~al.(2017)Felbo, Mislove, S{\o}gaard, Rahwan, and
  Lehmann}]{felbo2017}
Bjarke Felbo, Alan Mislove, Anders S{\o}gaard, Iyad Rahwan, and Sune Lehmann.
  2017.
\newblock Using millions of emoji occurrences to learn any-domain
  representations for detecting sentiment, emotion and sarcasm.
\newblock In \emph{Conference on Empirical Methods in Natural Language
  Processing (EMNLP)}.

\bibitem[{Huang et~al.(2018)Huang, Zaiane, Trabelsi, and Dziri}]{NAACL18}
Chenyang Huang, Osmar~R. Zaiane, Amine Trabelsi, and Nouha Dziri. 2018.
\newblock Automatic dialogue generation with expressed emotions.
\newblock In \emph{16th Annual Conference of the North American Chapter of the
  Association for Computational Linguistics (NAACL)}, New Orleans, USA.

\bibitem[{Kiritchenko et~al.(2014)Kiritchenko, Zhu, and
  Mohammad}]{kiritchenko2014sentiment}
Svetlana Kiritchenko, Xiaodan Zhu, and Saif~M Mohammad. 2014.
\newblock Sentiment analysis of short informal texts.
\newblock \emph{Journal of Artificial Intelligence Research}, 50:723--762.

\bibitem[{Lin et~al.(2017)Lin, Feng, Santos, Yu, Xiang, Zhou, and
  Bengio}]{lin2017structured}
Zhouhan Lin, Minwei Feng, Cicero Nogueira~dos Santos, Mo~Yu, Bing Xiang, Bowen
  Zhou, and Yoshua Bengio. 2017.
\newblock A structured self-attentive sentence embedding.
\newblock \emph{arXiv preprint arXiv:1703.03130}.

\bibitem[{Mohammad et~al.(2018)Mohammad, Bravo-Marquez, Salameh, and
  Kiritchenko}]{Mohammad-semeval18}
Saif Mohammad, Felipe Bravo-Marquez, Mohammad Salameh, and Svetlana
  Kiritchenko. 2018.
\newblock \href {https://doi.org/10.18653/v1/S18-1001} {Semeval-2018 task 1:
  Affect in tweets}.
\newblock In \emph{Proceedings of The 12th International Workshop on Semantic
  Evaluation}, pages 1--17. Association for Computational Linguistics.

\bibitem[{Mohammad and Turney(2013)}]{mohammad2013crowdsourcing}
Saif~M Mohammad and Peter~D Turney. 2013.
\newblock Crowdsourcing a word--emotion association lexicon.
\newblock \emph{Computational Intelligence}, 29(3):436--465.

\bibitem[{Pennington et~al.(2014)Pennington, Socher, and
  Manning}]{pennington2014glove}
Jeffrey Pennington, Richard Socher, and Christopher~D. Manning. 2014.
\newblock Glove: Global vectors for word representation.
\newblock In \emph{Empirical Methods in Natural Language Processing (EMNLP)},
  pages 1532--1543.

\bibitem[{Peters et~al.(2018)Peters, Neumann, Iyyer, Gardner, Clark, Lee, and
  Zettlemoyer}]{peters2018deep}
Matthew~E Peters, Mark Neumann, Mohit Iyyer, Matt Gardner, Christopher Clark,
  Kenton Lee, and Luke Zettlemoyer. 2018.
\newblock Deep contextualized word representations.
\newblock \emph{arXiv preprint arXiv:1802.05365}.

\bibitem[{Plutchik(2001)}]{plutchik2001}
Robert Plutchik. 2001.
\newblock The nature of emotions: Human emotions have deep evolutionary roots,
  a fact that may explain their complexity and provide tools for clinical
  practice.
\newblock \emph{American Scientist}, 89(4):344--350.

\bibitem[{Serban et~al.(2016)Serban, Sordoni, Bengio, Courville, and
  Pineau}]{serban2016building}
Iulian~V Serban, Alessandro Sordoni, Yoshua Bengio, Aaron Courville, and Joelle
  Pineau. 2016.
\newblock Building end-to-end dialogue systems using generative hierarchical
  neural network models.
\newblock In \emph{Thirtieth AAAI Conference on Artificial Intelligence}.

\bibitem[{Sordoni et~al.(2015)Sordoni, Bengio, Vahabi, Lioma, Grue~Simonsen,
  and Nie}]{sordoni2015hierarchical}
Alessandro Sordoni, Yoshua Bengio, Hossein Vahabi, Christina Lioma, Jakob
  Grue~Simonsen, and Jian-Yun Nie. 2015.
\newblock A hierarchical recurrent encoder-decoder for generative context-aware
  query suggestion.
\newblock In \emph{Proceedings of the 24th ACM International on Conference on
  Information and Knowledge Management}, pages 553--562. ACM.

\bibitem[{Sugiyama and Kawanabe(2012)}]{sugiyama2012machine}
Masashi Sugiyama and Motoaki Kawanabe. 2012.
\newblock \emph{Machine learning in non-stationary environments: Introduction
  to covariate shift adaptation}.
\newblock MIT press.

\bibitem[{Vaswani et~al.(2017)Vaswani, Shazeer, Parmar, Uszkoreit, Jones,
  Gomez, Kaiser, and Polosukhin}]{vaswani2017attention}
Ashish Vaswani, Noam Shazeer, Niki Parmar, Jakob Uszkoreit, Llion Jones,
  Aidan~N Gomez, {\L}ukasz Kaiser, and Illia Polosukhin. 2017.
\newblock Attention is all you need.
\newblock In \emph{Advances in Neural Information Processing Systems}, pages
  5998--6008.

\bibitem[{Yaddolahi et~al.(2017)Yaddolahi, Shahraki, and
  Zaiane}]{SentimentSurvey17}
Ali Yaddolahi, Ameneh~Gholipour Shahraki, and Osmar~R. Zaiane. 2017.
\newblock Current state of text sentiment analysis from opinion to emotion
  mining.
\newblock \emph{ACM Computing Surveys}, 50(2):25:1--25:33.

\end{thebibliography}
\bibliographystyle{acl_natbib}

\end{document}